\documentclass{article}
\usepackage[nonatbib]{neurips_2025}

\usepackage[numbers]{natbib}
\usepackage[utf8]{inputenc} 
\usepackage[T1]{fontenc}    
\usepackage{hyperref}       
\usepackage{url}            
\usepackage{booktabs}       
\usepackage{amsfonts}       
\usepackage{nicefrac}       
\usepackage{microtype}      
\usepackage{graphicx}
\usepackage{enumitem}
\usepackage{subcaption}
\usepackage{amsmath}
\usepackage{multirow}
\usepackage{bbm}
\usepackage{lineno}
\usepackage{cleveref}
\usepackage{graphicx}
\usepackage{placeins}
\usepackage[table]{xcolor}
\usepackage{wrapfig}
\usepackage{booktabs}
\usepackage{adjustbox}
\usepackage{color} 
\usepackage{xcolor} 
\definecolor{uiucdarkblue}{cmyk}{1, 0.65, 0, 0.2} 
\definecolor{adobered}{cmyk}{0, 1, 1, 0.1}        
\definecolor{nvidiagreen}{cmyk}{0.91, 0, 0.76, 0.46} 
\definecolor{linkblue}{RGB}{0, 102, 204} 
\definecolor{grayurl}{RGB}{80, 80, 80}
\usepackage{color}
\usepackage{hyperref}
\hypersetup{
  colorlinks=true,
  urlcolor=blue,  
}

\usepackage{footmisc} 

\newcommand{\frm}{FRAME}
\newcommand{\boldheader}[1]{\noindent\textbf{#1.}}

\definecolor{blue}{RGB}{200, 230, 242}

\title{\frm{}: Pre-Training Video \underline{F}eature \underline{R}epresentations via \underline{A}nticipation and \underline{Me}mory}

\author{
\textbf{Sethuraman T V}$^{1,2}$, \textbf{Savya Khosla}$^{2}$, \textbf{Vignesh Srinivasakumar}$^{2\dagger}$, 
\textbf{Jiahui Huang}$^{1}$, \textbf{Seoung Wug Oh}$^{1}$,\\ 
\textbf{Simon Jenni}$^{1}$, \textbf{Derek Hoiem}$^{2}$, \textbf{Joon-Young Lee}$^{1}$ \\
\hspace{-35pt}$^{1}$\textcolor{adobered}{\textbf{Adobe Research}}, 
$^{2}$\textcolor{uiucdarkblue}{\textbf{University of Illinois Urbana-Champaign}} \\
\small{\hspace{-35pt}$^\dagger$Now at \textcolor{nvidiagreen}{\textbf{NVIDIA}}}\\
\small{\hspace{-35pt}\textcolor{linkblue}{\texttt{Project: https://video-frame-encoder.github.io/vid-frame-encoder/}}}\\
}

\makeatletter
\@submissionfalse
\makeatother

\begin{document}

\maketitle

\begin{abstract}
Dense video prediction tasks, such as object tracking and semantic segmentation, require video encoders that generate temporally consistent, spatially dense features for every frame. However, existing approaches fall short: image encoders like DINO or CLIP lack temporal awareness, while video models such as VideoMAE underperform compared to image encoders on dense prediction tasks. We address this gap with \frm{}, a self-supervised video frame encoder tailored for dense video understanding. \frm{} learns to predict current and future DINO patch features from past and present RGB frames, leading to spatially precise and temporally coherent representations. To our knowledge, \frm{} is the first video encoder to leverage image-based models for dense prediction while outperforming them on tasks requiring fine-grained visual correspondence. As an auxiliary capability, \frm{} aligns its class token with CLIP’s semantic space, supporting language-driven tasks such as video classification. We evaluate \frm{} across \textit{six dense prediction tasks} on \textit{seven datasets}, where it consistently outperforms image encoders and existing self-supervised video models. Despite its versatility, \frm{} maintains a compact architecture suitable for a range of downstream applications.
\end{abstract}


\label{sec:intro}
\begin{figure}[ht]
    \centering
    \includegraphics[width=\linewidth]{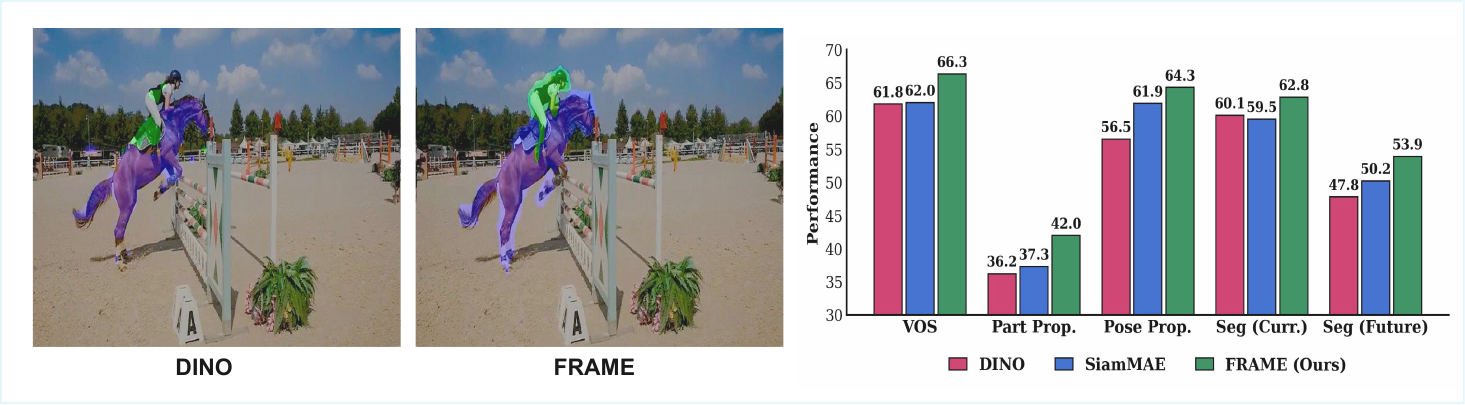}
        \caption{FRAME outperforms state-of-the-art self-supervised models (DINO, SiamMAE) on multiple dense video tasks. The student (FRAME) surpasses the teacher (DINO) by learning to predict current and future features using memory, improving temporal consistency and visual correspondence. (Right) Eg: VOS where FRAME improves segmentation of horse and rider. (left)" Tasks shown: VOS = Video Object Segmentation~\cite{davis}, Part Prop. = Part Propagation~\cite{vip}, Pose Prop. = Pose Propagation~\cite{jhmdb}, Seg = Semantic Segmentation~\cite{camvid} of current \& future frame.}
    \label{fig:final-teaser}
    \vspace{-10pt}
\end{figure}
\section{Introduction}
Our goal is to build a versatile \textbf{self-supervised video frame encoder for dense prediction tasks}. While self-supervised learning has achieved strong results in static image understanding and some video understanding tasks, it has yet to succeed in video dense prediction. Thus, we focus on tasks such as object tracking and semantic segmentation that require temporally consistent, spatially dense features for each frame. So far, progress in computer vision has been largely driven by static image encoders. Recent efforts have extended self-supervised learning to videos through masked autoencoding~\cite{cropmae,siamese_mae,dino_tracker,videomae} and contrastive learning~\cite{vifi_clip,vfs}. However, image encoders often outperform video-specific models across various video tasks~\cite{Aydemir2023SelfsupervisedOL,Qian2023SemanticsMT,Salehi2023TimeDT,Wang2022TokenCutSO,vificlip,actionclip,seeingflow}, especially on dense prediction tasks (see Table~\ref{label_prop_performance_table}). One explanation is that many video encoders are designed for global representation learning, such as classification, and fail to produce spatially detailed, temporally aligned frame-level features needed for dense prediction. Compounding this issue is the limited diversity and scale of video pretraining datasets compared to their image counterparts. Without such datasets, creating video frame encoders that can effectively capture scene dynamics, remember the past, interpret the present, and anticipate the future remains a significant challenge. 

\begin{figure*}[t]
    \centering
    \includegraphics[width=0.95\textwidth]{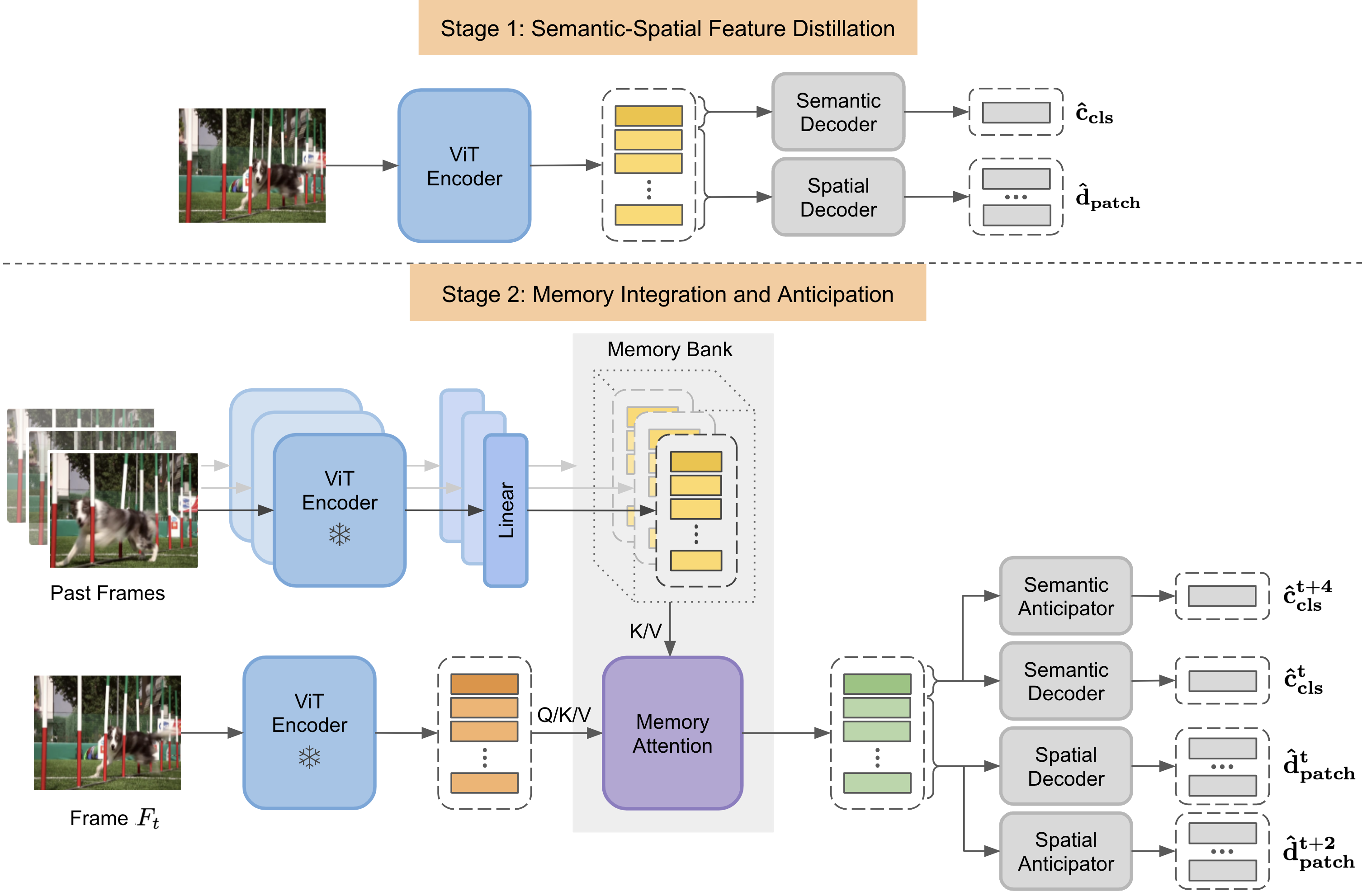} 
    \captionsetup{skip=2pt}
    \caption{{\bf Overview of \frm{} Architecture and Two-Stage Training Process.} In Stage 1, the encoder is trained to jointly distill CLIP features (providing semantic understanding) and DINO features (providing spatial understanding). In Stage 2, this pre-trained encoder processes the past and current frames and the model is optimized for memory integration and future anticipation.}
    \label{fig:framework}
\end{figure*}

To address this, we propose the \frm{} (\underline{F}eature \underline{R}epresentation and \underline{A}nticipation with \underline{ME}mory) encoder—a self-supervised video model that combines the strengths of pretrained image encoders with lightweight temporal modeling. Our key insight is that rich visual knowledge from large-scale image models (e.g., DINO, CLIP) can be transferred via feature distillation, avoiding the need for expensive video pretraining from scratch. At the same time, \frm{} remains compact and efficient, rather than building on top of heavy image backbones. As shown in Figure~\ref{fig:framework}, \frm{} is trained in two stages. In Stage 1, we train a student encoder to match dense patch-level and class-level features from frozen image-based teacher models (DINO and CLIP). In Stage 2, we equip the student with lightweight temporal modules—a memory unit that aggregates past context and an anticipation unit that predicts future features. This two-stage design preserves the spatial fidelity of image representations while introducing temporal consistency needed for dense video tasks. 

To our knowledge \frm{} is the first student video encoder distilled from image teachers that outperforms both the original image-only models (e.g., DINO, DINOv2) and prior self-supervised video encoders on frame-level dense prediction tasks such as semantic segmentation and label propagation. Additionally, we also align \frm{}'s class token with CLIP’s semantic space, enabling compatibility with language-driven objectives with minimal overhead. We validate this with video action classification as an initial baseline. \textit{Our results suggest that distilling image knowledge with lightweight temporal adaptation could be a scalable path for learning high-quality video representations in a self-supervised manner.}

\textbf{In summary, our main contributions are:}
\begin{itemize}[left=3pt, topsep=1pt, itemsep=1pt]
    \item \textbf{Model Architecture:} We propose \frm{}, a self-supervised video frame encoder that distills dense spatial and semantic features from pretrained image models (e.g., DINO, DINOv2, CLIP), and extends them into effective video encoders using lightweight memory and anticipation modules. The resulting model(s) produces temporally consistent per-frame representations and offers a practical foundation for dense video prediction tasks.

    \item \textbf{Performance Advancement:} \frm{} achieves state-of-the-art performance on dense video tasks—including semantic segmentation, label propagation, and visual correspondence—surpassing both image-only encoders and prior self-supervised video models.

    \item \textbf{Empirical Insights:} We conduct an extensive study of design factors, covering architectural and training choices such as encoder/decoder depth, input resolution, training duration, data size, and stage-wise training strategy. Code, model checkpoints, and training recipes will be released.
\end{itemize}

\section{Related Works}
\label{sec:related-works}
\vspace*{-1em}
Our approach integrates three key research areas: self-supervised visual representation learning, adaptation of image representations to video domains, and learning visual correspondence in videos.

\boldheader{Self-supervised Learning in Vision}
Self-supervised learning (SSL) has emerged as a powerful framework for learning visual representations without labels. Early methods relied on proxy tasks like patch ordering~\cite{doersch2015unsupervised} or frame shuffling~\cite{misra2016shuffle}, while more recent work focuses on contrastive learning~\cite{he2020momentum}, masked autoencoding~\cite{mae}, and self-distillation~\cite{dinov1,dinov2}. While masked autoencoders (e.g., MAE~\cite{mae}) excel at global representation learning, they struggle with dense prediction tasks. In contrast, DINO~\cite{dinov1,dinov2} achieves strong spatial correspondence and has become a preferred backbone for tasks like segmentation and tracking. Surprisingly, despite being trained only on images, DINO and similar image-based models often outperform video-specific self-supervised methods on dense prediction tasks (Table~\ref{label_prop_performance_table}). Even video adaptations such as VideoMAE~\cite{videomae} fail to close this gap. This performance gap motivates our approach:  rather than relying solely on large-scale video pretraining, we distill spatial and semantic features from pretrained image models, and enhance them with lightweight temporal modules for dense prediction. 

\boldheader{Extending Image Representations to Video}
Recent work has explored adapting image-based self-supervised methods to video tasks by modeling temporal relationships. SiamMAE~\cite{siamese_mae} and CropMAE~\cite{cropmae} extend masked image modeling to videos, but remain limited to pairwise frame relationships and require re-learning representations from scratch. DINO-Tracker~\cite{dino_tracker} instead fine-tunes pre-trained DINO~\cite{dinov1} for video correspondence, while contrastive video methods~\cite{vfs,vfs_2,vfs_3} depend on carefully tuned augmentations~\cite{what_should_not_be_in_contrastive_learning} and collapse-prevention mechanisms~\cite{add_1,add_2,add_3}. Other works use frozen image encoders and introduce auxiliary modules for temporal modeling~\cite{ranasinghe2022self,dave2024no}, but often fail to target video dense prediction. In contrast, \frm{} distills spatial features from DINO and semantic features from CLIP into a per-frame encoder without retraining or fine-tuning. While recent work explores similar distillation strategies for images~\cite{ranzinger2024radio,sameni2024building}, we demonstrate their effectiveness in video. Our method goes beyond pairwise modeling by incorporating memory and anticipation modules, yielding temporally consistent and predictive features critical for dense video prediction. \newline

\vspace{-10pt}

\boldheader{Visual Correspondence Learning}
Visual correspondence spans fine-grained (pixel/part) and object-level tracking, supporting tasks like video object segmentation~\cite{davis,youtubevos}, part propagation, and pose tracking~\cite{vip,jhmdb}. Traditional methods rely on supervision~\cite{vos_1,vos_2,cutie,deva,video_flow_tracking,video_semantic_segmentation}, limiting scalability due to annotation cost. While self-supervised learning has matched supervised performance in static images~\cite{dinov1,dinov2}, similar success in videos—especially for dense, spatio-temporal tasks—remains limited. Our approach addresses this by leveraging strong image encoders and introducing memory-based temporal modeling, enabling correspondence learning without labels.

\section{\frm{}}
\vspace*{-1em}
\vspace{3pt}
Our goal is to train a video frame encoder that captures spatial and semantic details from the current frame, enriched with temporal context from the past and anticipatory cues for the future. To this end, we adopt a two-stage training approach (Figure~\ref{fig:framework}): Stage 1 (Section~\ref{sec:stage-1}) distills semantic and spatial features from individual frames, yielding a strong strong frame encoder; Stage 2 (Section~\ref{sec:stage-2}) infuses the ability to capture past context and anticipate the near future, enabling it to generate temporally consistent frame-wise features.

\subsection{Stage 1: Semantic-Spatial Feature Distillation}
\label{sec:stage-1}
We train a ViT-based encoder with two lightweight decoders: one to match CLIP’s \texttt{[CLS]} token and another to match DINO’s patch-level features. This stage produces a per-frame encoder with rich semantic and spatial representations.

\boldheader{ViT Encoder}
We the input image $\mathbf{x} \in \mathbb{R}^{H \times W \times 3}$, into $N = HW / P^2$ non-overlapping patches and project each to a $D$-dimensional embedding, where $D$ is 384 or 768 depending on the backbone (e.g., DINO-S vs DINO-B/DINOv2). These are concatenated with a learnable $\texttt{[CLS]}$ token and positional embeddings to form the sequence $\mathbf{z_0} \in \mathbb{R}^{(N+1) \times D}$. A 12-layer ViT encoder processes this sequence, producing a $\texttt{[CLS]}$ token for global image representation $\mathbf{y_{cls}} \in \mathbb{R}^{1 \times D}$ and contextualized patch tokens for local image features $\mathbf{y_{patch}} \in \mathbb{R}^{N \times D}$.

\boldheader{Semantic Decoder}
The semantic decoder is a single linear layer that maps the ViT Encoder's $\texttt{[CLS]}$ token representation into CLIP's feature space. Formally, it projects $\mathbf{y_{cls}} \in \mathbb{R}^{1 \times D}$ to $\mathbf{\hat{c}_{cls}} \in \mathbb{R}^{1 \times D^c}$, where $D^c$ denotes CLIP's feature dimension size.

\boldheader{Spatial Decoder}
The spatial decoder is a single block Transformer that operates on the $N$ patch tokens generated by the ViT to project them into DINO's feature space. Formally, it processes $\mathbf{y_{patch}} \in \mathbb{R}^{N \times D}$ to produce $\mathbf{\hat{d}_{patch}} \in \mathbb{R}^{N \times D^d}$, where $D^d$ denotes DINO's feature dimension size.

\boldheader{Training}
The encoder and the decoders are jointly trained in an end-to-end manner using the following loss function:
\begin{equation}
    \mathcal{L} = \lambda_1 \Big( 1- \frac{\mathbf{c_{cls}} \cdot \mathbf{\hat{c}_{cls}}}{\|\mathbf{c_{cls}}\| \|\mathbf{\hat{c}_{cls}}\|} \Big) + \lambda_2  \frac{\sum_{\text{patch}}||\mathbf{d_{patch}} - \mathbf{\hat{d}_{patch}}||^2}{N}
\end{equation}
We use $\lambda_1 = 1.0$ and $\lambda_2 = 1.0$,  $\mathbf{c_{cls}}$ and $\mathbf{d_{patch}}$ refer to CLIP and DINO teacher outputs, respectively. Both the semantic and spatial decoders are intentionally kept lightweight to encourage the encoder to produce directly usable representations, rather than offloading learning to the decoders.

To ensure balanced learning, we normalize both loss components via mean scaling and apply cosine annealing with warm restarts to avoid convergence to local minima. These strategies improve class token alignment with CLIP. We also investigate loss dropout—temporarily disabling one loss component—and gradient-based loss balancing, with details provided in the supplementary material.

\boldheader{Inference}
Once training is complete, the decoders are discarded. The encoder is retained as a compact image backbone that produces both semantic and spatially rich features for each frame, which are used for encoding in the next stage.
\subsection{Stage 2: Memory Integration and Anticipation}
\label{sec:stage-2}
While Stage 1 produces rich features for individual frames, video tasks require representations that are temporally consistent and predictive. Stage 2 addresses this by introducing temporal context through a memory module that integrates past frame information and anticipates near-future representations—critical for dynamic agents where processing delays necessitate short-term forecasting. We freeze the ViT encoder from Stage 1 and use it to extract features from the current and past frames. These are fused via a memory block to produce temporally enriched representations for the current frame. Future frame features serve as a self-supervised signal to guide this integration.

\boldheader{Memory Block}
The memory block consists of two components: a memory bank and a memory attention module. The memory bank is a first-in-first-out (FIFO) queue that stores the encoding of the past frames, and the memory attention module is trained to enrich the current frame's representations with the context derived from past frames. 
When encoding a frame $F_t$ at timestep $t$, the memory bank stores encodings of previous $m=5$ frames using the frozen ViT encoder from Stage 1.  Each frame yields a set of patch tokens $\in \mathbb{R}^{N \times D}$, where $D$ is the feature dimension (e.g., 384 for DINO-S, 768 for DINO-B/DINOv2). To focus on essential information, each token is passed through a linear projection to a lower-dimensional space ($d = 64$), forming a compact representation that acts as a bottleneck to retain salient temporal features, particularly inspired by the memory attention block of SAM 2~\cite{sam2}. These reduced features are stored in a FIFO memory bank, which is continuously updated as new frames are processed. To encode positional context, we enrich these features with spatio-temporal positional embeddings that encode both patch location and frame timestamp.

\boldheader{Memory Attention Module}
To integrate information from memory, we concatenate all stored frame features along the token dimension and apply a linear projection to produce consolidated memory embeddings. These are fed into a cross-attention module alongside the current frame’s features. Here, queries come from the current frame, while keys and values are derived from the memory bank and the current frame itself. This mechanism enables the encoder to dynamically attend to relevant past information when computing the current representation. To further refine the output, we follow the cross-attention with a self-attention block applied over the updated current frame features. The resulting features from this module serve as temporally enriched representations that support downstream tasks requiring frame-to-frame consistency and short-term prediction.

\boldheader{Semantic and Spatial Decoders}
To train the memory-augmented encoder to model temporal consistency and short-term anticipation, we use four lightweight decoders: (1) A {\em semantic decoder} to predict CLIP-like global features $\mathbf{\hat{c}_{cls}^{t}}$ for the current frame $F_{t}$, (2) a {\em semantic anticipator} to anticipate CLIP-like global features $\mathbf{\hat{c}_{cls}^{t+4}}$ for the future frame $F_{t+4}$, (3) a {\em spatial decoder} to predict DINO-like patch-level features $\mathbf{\hat{d}_{patch}^{t}}$ for the current frame $F_{t}$, and (4) a {\em spatial anticipator} to anticipate DINO-like patch-level features $\mathbf{\hat{d}_{patch}^{t+2}}$ for the future frame $F_{t+2}$. These decoders force the current frame representation to encode predictive signals for future frames. The selection of the future frame to anticipate is informed by an experiment measuring the variability of CLIP and DINO features across consecutive video frames (details provided in the supplementary). We choose a frame delta that demonstrates moderate changes -- enough for meaningful learning without overwhelming the model. In our analysis, CLIP features exhibit slower variation due to the gradual change of global information, while DINO features fluctuate more rapidly due to their sensitivity to local patch-level changes. 

\boldheader{Training}
We train the memory attention module, linear projection layers, and decoders using the following loss:
\begin{equation}
\begin{aligned}
    \mathcal{L} = & \alpha_1 \Big( 1 - \frac{\mathbf{c_{cls}^{t}} \cdot \mathbf{\hat{c}_{cls}^{t}}}{\|\mathbf{c_{cls}^{t}}\| \|\mathbf{\hat{c}_{cls}^{t}}\|} \Big) + 
    \alpha_2 \Big( 1 - \frac{\mathbf{c_{cls}^{t+4}} \cdot \mathbf{\hat{c}_{cls}^{t+4}}}{\|\mathbf{c_{cls}^{t+4}}\| \|\mathbf{\hat{c}_{cls}^{t+4}}\|} \Big)
    + \alpha_3 \frac{\sum_{\text{patch}} ||\mathbf{d_{patch}^{t}} - \mathbf{\hat{d}_{patch}^{t}}||^2}{N} \\
    & + \alpha_4 \frac{\sum_{\text{patch}} ||\mathbf{d_{patch}^{t+2}} - \mathbf{\hat{d}_{patch}^{t+2}}||^2}{N}
\end{aligned}
\end{equation}
Here, $\alpha_1=0.2$, $\alpha_2=0.1$, $\alpha_3=2.0$, $\alpha_4=0.4$, and $\mathbf{c_{cls}^{i}}$ and $\mathbf{d_{patch}^{i}}$ are the CLIP and DINO features of $i^{th}$ frame. These weights ($\alpha_1$, $\alpha_2$, $\alpha_3$ , $\alpha_4$) were empirically determined based on results from a subset of DAVIS \cite{davis} dataset.

\subsection{Task-Specific Feature Adaptation}
\begin{figure*}[t]
    \centering
    \begin{subfigure}[t]{0.23\textwidth}
        \centering
        \includegraphics[width=\textwidth]{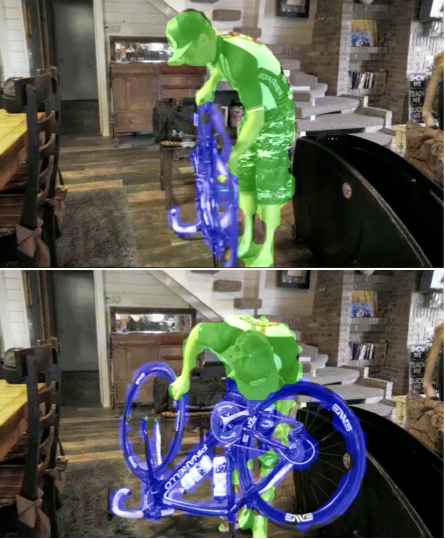}
        \caption{Video Object Segmentation}
        \label{fig:figure1}
    \end{subfigure}
    \hfill
    \begin{subfigure}[t]{0.23\textwidth}
        \centering
        \includegraphics[width=\textwidth]{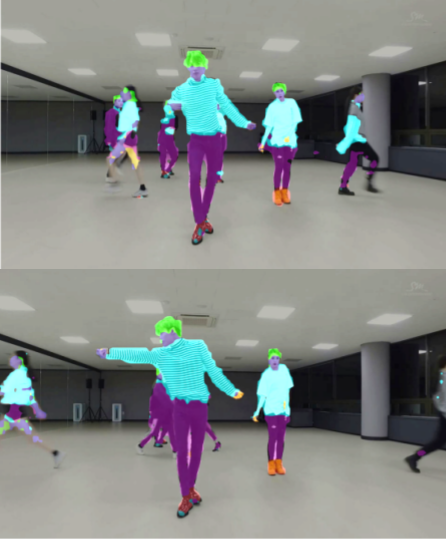}
        \caption{Semantic Part Propogation}
        \label{fig:figure2}
    \end{subfigure}
    \hfill
    \begin{subfigure}[t]{0.23\textwidth}
        \centering
        \includegraphics[width=\textwidth]{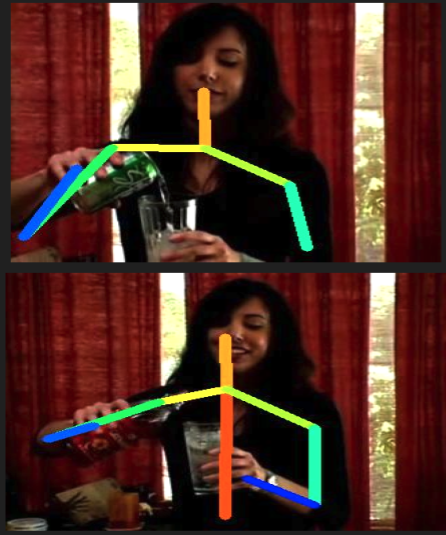}
        \caption{Pose Propagation}
        \label{fig:figure3}
    \end{subfigure}
     \hfill
    \begin{subfigure}[t]{0.23\textwidth}
        \centering
        \includegraphics[width=\textwidth]{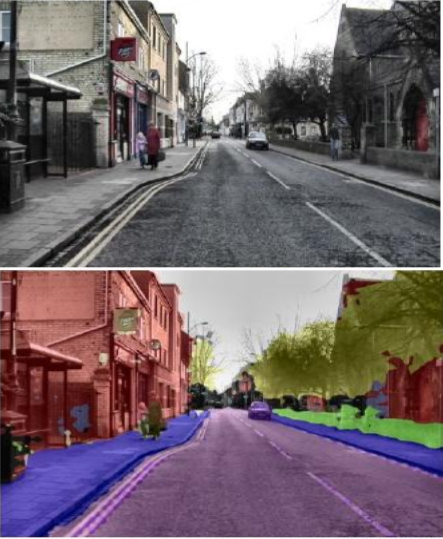}
        \caption{Semantic Segmentation}
        \label{fig:figure4}
    \end{subfigure}
    \caption{{\bf Examples of correspondence and segmentation tasks} (a) Video object segmentation: initial (top) and propagated (bottom) frames. (b) Semantic part propagation: initial (top) and propagated (bottom) frames. (c) Pose propagation: initial (top) and propagated (bottom) poses. (d) Semantic segmentation: original image (top) and segmentation overlay (bottom).}
    \label{fig:three_figures}
    \vspace{-12pt}
\end{figure*}

At inference time, we discard all decoders and use only the patch and \texttt{[CLS]} token outputs from the memory attention module (Figure~\ref{fig:framework})for downstream tasks~\ref{fig:figure4}. For tasks requiring fine-grained spatial-temporal reasoning—such as label propagation, semantic segmentation, and pose tracking—we use the patch tokens. For tasks requiring global frame-level understanding—such as video action classification—we use the \texttt{[CLS]} token. For zero-shot classification aligned with CLIP (e.g., prompt-based action recognition), we project the \texttt{[CLS]} token into CLIP space using the semantic decoder $\mathbf{\hat{c}_{cls}^{t}}$ of the current frame. This dual-mode design enables \frm{} to support both zero-shot and trainable downstream tasks without modifying the encoder backbone.

\section{Experiments}
\begin{table*}[ht]
\centering
\caption{\textbf{Semi-supervised visual correspondence task comparison.} We compare performance on DAVIS~\cite{davis}, VIP~\cite{vip}, and JHMDB~\cite{jhmdb} datasets where \frm{} substantially
outperforms. }
{\tiny
\setlength{\tabcolsep}{1.3pt} 
\renewcommand{\arraystretch}{1.0} 
\resizebox{\linewidth}{!}{ 
\begin{tabular}{c c c c ccc cc}
\toprule
\multirow{2}{*}{\textbf{Method}} & \multirow{2}{*}{\textbf{Backbone}} & \multirow{2}{*}{\textbf{Dataset}} & \multirow{2}{*}{\textbf{Type}} & \multicolumn{3}{c}{\textbf{DAVIS}} & \textbf{VIP} & \textbf{JHMDB} \\
\cmidrule(lr){5-7} \cmidrule(lr){8-8} \cmidrule(lr){9-9}
& & & & $J\&F_m$ & $J_m$ & $F_m$ & mIoU & PCK@0.1 / PCK@0.2 \\
\midrule
SimSiam~\cite{chen2020simsiam} & ResNet-50 & ImageNet & Image & 66.3 & 64.5 & 68.2 & 35.0 & 58.4 / 77.5 \\
MoCo~\cite{he2019moco} & ResNet-50 & ImageNet & Image & 65.4 & 63.2 & 67.6 & 36.1 & 60.4 / 79.3 \\
\midrule
TimeCycle~\cite{CVPR2019_CycleTime} & ResNet-50 & VLOG & Video & 40.7 & 41.9 & 39.4 & 28.9 & 57.7 / 78.5 \\
UVC~\cite{uvc_2019} & ResNet-50 & Kinetics & Video & 56.3 & 54.5 & 58.1 & 34.2 & 56.0 / 76.6 \\
VFS~\cite{vfs} & ResNet-50 & Kinetics & Video & 68.9 & 66.5 & 71.3 & 43.2 & 60.9 / 80.7 \\
\midrule
MAE~\cite{mae} & ViT-B/16 & ImageNet & Image & 53.5 & 52.1 & 55.0 & 28.1 & 44.6 / 73.4 \\
VideoMAE~\cite{videomae} & ViT-S/16 & Kinetics & Video & 39.3 & 39.7 & 38.9 & 23.3 & 41.0 / 67.9 \\
DINO~\cite{dinov1} & ViT-S/16 & ImageNet & Image & 61.8 & 60.2 & 63.4 & 36.2 & 45.6 / 75.0 \\
SiamMAE~\cite{siamese_mae} & ViT-S/16 & Kinetics & Video & 62.0 & 60.3 & 63.7 & 37.3 & 47.0 / 76.1 \\
\textbf{\frm{}} & \textbf{ViT-S/16} & \textbf{Kinetics} & \textbf{Video} & \textbf{65.7} & \textbf{62.1} & \textbf{69.2} & \textbf{41.2} & \textbf{48.7} / \textbf{79.2} \\
\textbf{\frm{}} & \textbf{ViT-S/16} & \textbf{Kinetics+Ego4D} & \textbf{Video} & \textbf{66.3} & \textbf{62.9} & \textbf{69.8} & \textbf{42.0} & \textbf{49.0} / \textbf{79.3} \\
\midrule
DINO~\cite{dinov1} & ViT-S/8 & ImageNet & Image & 69.9 & 66.6 & 73.1 & 39.5 & 56.5 / 80.3 \\
SiamMAE~\cite{siamese_mae} & ViT-S/8 & Kinetics & Video & 71.4 & 68.4 & 74.5 & 45.9 & 61.9 / 83.8 \\
\textbf{\frm{}} & \textbf{ViT-S/8} & \textbf{Kinetics} & \textbf{Video} & \textbf{73.2} & \textbf{69.5} & \textbf{77.0} & \textbf{47.9} & \textbf{64.1} / \textbf{85.9} \\
\textbf{\frm{}} & \textbf{ViT-S/8} & \textbf{Kinetics+Ego4D} & \textbf{Video} & \textbf{74.4} & \textbf{69.9} & \textbf{79.0} & \textbf{48.2} & \textbf{64.3} / \textbf{86.0} \\

\bottomrule
\end{tabular}
}
}
\label{label_prop_performance_table}
\end{table*}

We evaluate \frm{}'s representation on three categories of video tasks: (1) \textbf{visual correspondence}—including video object segmentation, semantic part propagation, and pose tracking—to assess the model's ability to learn fine-grained, temporally consistent correspondences; (2) \textbf{semantic segmentation}, to evaluate spatial precision and temporal coherence; and (3) \textbf{action classification}, to assess global frame-level action categorization.

\boldheader{Training details} As a self-supervised method, \frm{} can be trained on any video data. We primarily train on 80,000 videos from Kinetics-400~\cite{kinetics} (uniformly sampled per class) and 700 randomly sampled videos from Ego4D~\cite{Grauman2021Ego4DAT}. We repeat training with four different Ego4D subsets to measure variability and also report results using the full Ego4D and Kinetics dataset in the supplementary. We found that performance remained largely stable after a certain threshold of training videos and epochs. All models are trained for 70 epochs using $400 \times 400$ resized frames. We report results from two main variants: one trained on Kinetics only, and one trained on Kinetics + Ego4D, with the latter consistently outperforming due to increased data diversity. An extended data ablation—including results with SA-V~\cite{sam2}—is provided in the supplementary.

\begin{figure}[t]
    \centering
    \includegraphics[width=\linewidth]{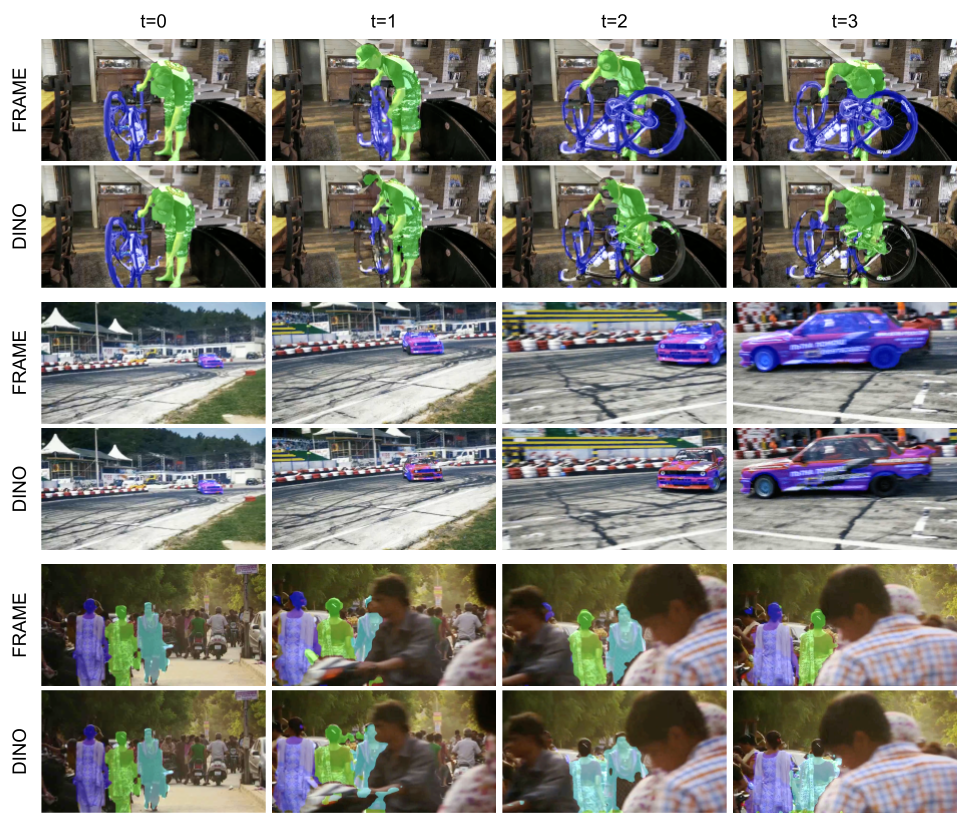}
    \caption{{\bf Comparison of \frm{} and DINO on feature propagation across video frames.} \frm{} demonstrates greater robustness to viewpoint changes, occlusions, and object reappearances, making it a more suitable video frame encoder.
}
    \label{fig:qualitative_examples}
    \vspace{-12pt}
\end{figure}

\boldheader{Visual Correspondence Tasks}
To assess the \frm{} encoder's feature quality, we use three semi-supervised video propagation tasks: object segmentation on DAVIS-2017~\cite{davis}, semantic part tracking on VIP~\cite{vip}, and pose tracking on JHMDB~\cite{jhmdb}. Starting with a ground truth mask for the first frame, we propagate it through the video using patchwise k-nearest neighbor matching from each frame to the next. We use this simple propagation technique to isolate the effectiveness of \frm{} to produce temporally consistent features and to enable direct comparison with previous pretrained encoders. This evaluation is not intended to showcase the best possible performance via task-specific tuning. Instead, it aims to assess feature quality under a standard setting consistent with the literature \cite{spacetimecorres, siamese_mae, vfs, cropmae}. Our results for these tasks are in Table~\ref{label_prop_performance_table}, with results for some comparison methods drawn from a similar table in~\cite{siamese_mae}. In the supplemental, we provide videos comparing FRAME and DINO, along with an extended label propagation table that highlights: (1) performance gains from ViT-S to ViT-B across tasks, (2) gains over DINOv1 and DINOv2 across different patch sizes, and (3) runtime, parameter count, and memory usage details.


\begin{wraptable}[11]{r}{0.55\textwidth} 
    \vspace{-1pt}
     \captionof{table}{\textbf{Video activity classification comparison.} \frm{} performs comparably to CLIP in both zero-shot and linear classification.}
    \label{tab:action_classification_comparison}
    \setlength{\tabcolsep}{1.7pt}
    \begin{adjustbox}{max width=\linewidth}
        \begin{tabular}{lcccc}
            \toprule
            \textbf{Model} & \multicolumn{2}{c}{\textbf{Zero-Shot (\%)}} & \multicolumn{2}{c}{\textbf{Linear (\%)}} \\
             & HMDB-51 & UCF-101 & HMDB-51 & UCF-101 \\
            \midrule
            CLIP & \textbf{52.7} & \textbf{59.1} & 76.2 & 69.5 \\
            FRAME S/16 & 52.1 & 58.5 & 76.1 & \textbf{69.7} \\
            FRAME B/16 & 52.4 & 58.9 & \textbf{77.5} & 69.3 \\
            \bottomrule
        \end{tabular}
    \end{adjustbox}
\end{wraptable}

\begin{figure}[t]
    \centering
    \begin{minipage}[t]{0.44\textwidth}
        \vspace{0pt}
        \centering
        \captionof{table}{\textbf{Performance on DAVIS ($J\&F_m$).} \frm{} uses region-based representations~\cite{dinosam} to track objects from a single annotated frame, outperforming DINO variants and approaching task-specific models like SAM 2.(details in appendix)}
        \label{tab:davis_region_results}
        \scriptsize
        \setlength{\tabcolsep}{2pt}
        \renewcommand{\arraystretch}{0.98}
        \resizebox{0.8\linewidth}{!}{%
        \begin{tabular}{lc}
            \toprule
            \textbf{Model} & \textbf{DAVIS ($J\&F_m$)} \\
            \midrule
            SAM 2 & \textbf{90.7} \\
            DINO ViT-S/8 & 80.7 \\
            \rowcolor{cyan!15}
            FRAME ViT-S/8 & 84.4 \\
            DINO ViT-B/8 & 85.6 \\
            \rowcolor{cyan!15}
            FRAME ViT-B/8 & 88.1 \\
            \bottomrule
        \end{tabular}
        }
    \end{minipage}
    \hfill
    \begin{minipage}[t]{0.5\textwidth}
        \vspace{-3pt}
        \centering
        \includegraphics[width=\linewidth, height=4cm, keepaspectratio=false]{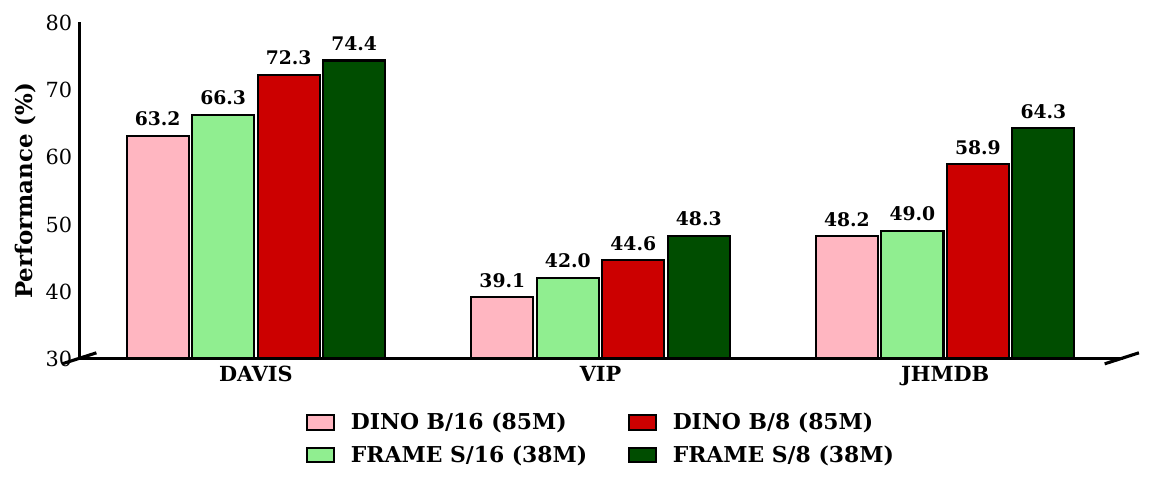}
        \captionof{figure}{\textbf{Comparison of \frm{} and DINO across model scales.} \frm{} outperforms DINO with fewer parameters.}
        \label{fig:parameters_chart}
    \end{minipage}
    
\end{figure}

\boldheader{Video Object Segmentation}
We evaluate \frm{} on the DAVIS-2017~\cite{davis} benchmark for semi-supervised multi-object video segmentation. Consistent with previous methods \cite{siamese_mae, videowalk, cropmae}, we use 480p images and the standard evaluation setup for comparability. Performance is measured by the Jaccard Index, which assesses the overlap between the predicted and ground truth masks, and the boundary accuracy, which evaluates how closely the predicted mask aligns with the object’s boundaries. \frm{} outperforms SiamMAE~\cite{siamese_mae}, 65.7 vs.\ 62.0 with a patch size of 16, and 73.2 vs.\ 71.4 with a patch size of 8, indicating strong feature quality and consistency in segmentation (Table \ref{label_prop_performance_table}). While task-specific fine-tuned models still outperform\cite{cutie}, \frm{} significantly narrows the gap to these specialized methods—despite using simple label propagation and no task-specific tuning or learned memory modules for inference. As an auxiliary experiment to demonstrate that \frm{} can approach models trained on large-scale video segmentation data, we pool region tokens from \frm{} features using class-agnostic SAM\cite{SAM}  masks and track objects via cosine similarity to the first-frame token, following the region-based representation paradigm~\cite{dinosam}. Results in Table~\ref{tab:davis_region_results} show that \frm{} outperforms DINO variants and approaches SAM~2. These results suggest that FRAME + SAM~2 could enable a more memory-efficient many-object tracking. See supplementary for additional details of the experiment. 

\boldheader{Semantic Part Propagation}
Next, we evaluate \frm{} on the Video Instance Parsing (VIP)~\cite{vip} benchmark, a task that involves propagating semantic masks for 20 different human parts. The VIP dataset presents unique challenges compared to others in our evaluation due to its much longer video durations (up to 120 seconds). Following the protocol from prior work~\cite{eval_protocol_vip}, we use images at $560 \times 560$ resolution and rely on a single context frame for evaluation. On this challenging dataset, our ViT-S/16 model and ViT-S/8 variants outperform prior methods, with a mean Intersection-over-Union (mIoU) improvement over prior self-supervised SotA SiamMAE~\cite{siamese_mae}, from 37.3 to 41.2 and 45.9 to 47.9 respectively (Table \ref{label_prop_performance_table}). The mIoU metric measures the overlap between predicted and ground truth masks, capturing the model's accuracy in segmenting and tracking each human part. 


\boldheader{Human Part Propagation}
We evaluate \frm{} on the keypoint propagation task, which requires the accurate spatial tracking of 15 keypoints across frames. This task demands highly precise correspondence between frames to correctly propagate keypoint locations. Following the protocol from prior work~\cite{eval_protocol_vip}, we use $320 \times 320$ images and a single context frame for evaluation. \frm{} surpasses all previous methods in this task, with \frm{} (S/16) improving PCK@0.1 by 1.7\% and \frm{} (S/8) by 2.4\% over SiamMAE~\cite{siamese_mae} (Table \ref{label_prop_performance_table}). PCK@0.1 assesses keypoint accuracy within 10\% of the object's bounding box, while PCK@0.2 allows a 20\% margin. Higher PCK scores indicate better spatial correspondence and keypoint tracking precision.

\begin{figure}[t]
    \centering

    \begin{minipage}[t]{0.48\textwidth}
        \vspace{0pt}
        \centering
        \includegraphics[width=\linewidth]{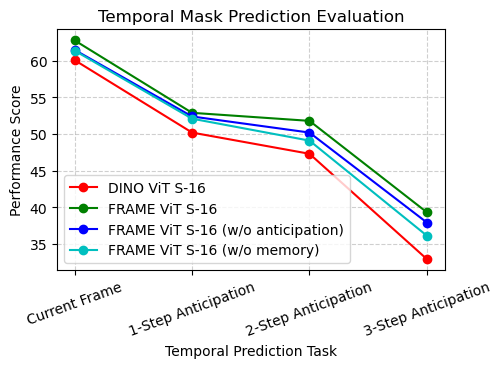}
        \vspace{-5pt}
        \captionof{figure}{\textbf{(a) Semantic segmentation on current and future frames.} \frm{} outperforms DINO on CamVid. Removing memory or anticipation reduces performance, showing their complementary role in temporal reasoning.}
        \label{fig:anticipation}
    \end{minipage}
    \hfill
    \begin{minipage}[t]{0.48\textwidth}
        \vspace{0pt}
        \centering
        \captionof{table}{\textbf{(b) Video semantic segmentation results on CamVid and VSPW.} \frm{} outperforms other self-supervised methods on both datasets and tasks. }
        \label{tab:segmentation_comparison_2}
        \footnotesize
        \setlength{\tabcolsep}{2pt}
        \renewcommand{\arraystretch}{1.1}
        \resizebox{\linewidth}{!}{%
            \begin{tabular}{lcccc}
                \toprule
                \multirow{2}{*}{\textbf{Model}} & \multicolumn{2}{c}{\textbf{CamVid (mIoU)}} & \multicolumn{2}{c}{\textbf{VSPW (mIoU)}} \\
                \cmidrule(lr){2-3} \cmidrule(lr){4-5}
                 & Curr. & Future & Curr. & Future \\
                \midrule
                DINO ViT-S/16 & 60.1 & 50.2 & 36.4 & 25.6 \\
                SiamMAE ViT-S/16 & 59.5 & 47.8 & 34.7 & 24.9 \\
                CatMAE ViT-S/16 & 58.9 & 46.2 & 35.1 & 25.1 \\
                CropMAE ViT-S/16 & 56.4 & 47.9 & 35.8 & 23.7 \\
                \textbf{\frm{} ViT-S/16} & \textbf{62.8} & \textbf{53.9} & \textbf{38.9} & \textbf{28.3} \\
                \midrule
                DINO ViT-S/8 & 59.6 & 51.1 & 35.9 & 25.8 \\
                CatMAE ViT-S/8 & 58.9 & 50.5 & 34.7 & 25.4 \\
                \textbf{\frm{} ViT-S/8} & \textbf{62.6} & \textbf{54.0} & \textbf{38.0} & \textbf{27.4} \\
                \midrule 
                DINOv2 ViT-L/14 & 68.3 & 56.1 & 41.8 & 30.3 \\
                \textbf{\frm{} ViT-L/14} & \textbf{69.8} & \textbf{59.2} & \textbf{44.0} & \textbf{33.8} \\
                \bottomrule
            \end{tabular}
        }
        \vspace{-5pt}
    \end{minipage}

\end{figure}

\boldheader{Video Semantic Segmentation}
We evaluate video semantic segmentation on CamVid~\cite{camvid} and VSPW~\cite{vspw}, covering urban and diverse real-world scenes. A linear decoder is trained atop DINO~\cite{dinov1}, DINOv2~\cite{dinov2}, and \frm{} features. As shown in Table~\ref{tab:segmentation_comparison_2}, \frm{} outperforms all prior self-supervised encoders across backbones. With ViT-S/8, \frm{} improves over DINO by 3.0\% (current) and 2.9\% (future) on CamVid, and by 2.8\% and 1.6\% on VSPW. With ViT-L/14, it surpasses DINOv2 by 1.5–3.5\% across tasks. When paired with a U-Net decoder (Table~\ref{tab:segmentation_comparison} (supplemental)), \frm{} achieves 79.1 mIoU on CamVid and 52.3 on VSPW—trailing task-specific state-of-the-art methods, but that is not our primary goal. Rather, our goal is to demonstrate that \frm{} provides a general-purpose dense representation that can approach or match SotA when paired with strong decoders, for eg. improving over DINOv2 by up to 4.8\%. Figure~\ref{fig:anticipation} extends this analysis to future-frame prediction on CamVid. \frm{} consistently outperforms DINO, with a growing advantage at $t+3$, surpassing its $t+2$ training objective. This highlights \frm{}'s ability to capture temporal structures, enhancing predictive robustness.

\boldheader{Video Action Classification}
To assess suitability for video action classification, we evaluate on HMDB-51~\cite{hmdb51} and UCF-101~\cite{ucf101} (Table~\ref{tab:action_classification_comparison}). For zero-shot evaluation, we sample eight frames, extract CLS tokens from either CLIP ViT-B/32~\cite{clip} or the \frm{} (S/16) semantic decoder, and average them into a $1 \times 512$ video representation. Cosine similarity with text-encoded action labels (e.g., “a person \textit{brushing}”) is used for prediction. \frm{} achieves performance comparable to CLIP, despite being trained only on Kinetics-400~\cite{kinetics}. In the supervised setting, a linear classifier on the \texttt{[CLS]} token yields similar results, confirming \frm{}’s versatility for classification, segmentation, and correspondence. Additional experiments distilling CLIP ViT-L/14 into \frm{} (S/16) and other variants using DINO and DINOv2 backbones are included in the supplementary.

\boldheader{Ablations} 
In Table~\ref{tab:ablation-study}, we analyze the impact of key components of the \frm{} encoder: the stagewise training, use of memory, anticipation loss, and training data. Memory and anticipation substantially boost the visual correspondence tasks. They do not help video clip action classification (HMDB, UCF) because all frames are available at once for that task.  Re-using and freezing the Stage 1 ViT Encoder for Stage 2 provides the best results, considering the added efficiency in training and that using the same ViT encoder for past and current frame encoding saves computation and memory. 
The inclusion of Ego4d videos in pretraining further boosts performance by adding diversity, with tasks like semantic segmentation on CamVid benefiting the most.

\FloatBarrier

\begin{table*}[ht]
\centering
\footnotesize
\captionsetup{skip=5pt}
\caption{
\textbf{Ablations.} Our method (blue highlight) is two-stage training, where Stage 1 trains the per-frame ViT Encoder, and then Stage 2 trains the Memory Attention module to anticipate using memory. We compare to using only the Stage 1 ViT Encoder (``Stage 1''),  training the stage 2 ViT random with initialization (``Stage 2 Scratch''), fine-tuning the ViT in the second stage (``2-Stage FT''), or re-using the frozen Stage 1 ViT in Stage 2 (``2-Stage''). We choose the last for efficiency and good results. Adding more data (+Ego4D) helps, and memory and learning to anticipate help for dense prediction tasks. 
}
\renewcommand{\arraystretch}{1.2}
\setlength{\tabcolsep}{2pt} 
\begin{tabular}{c c c c c c c c c c}
\hline
\multirow{2}{*}{\textbf{Memory}} & 
\multirow{2}{*}{\textbf{Anticipation}} & 
\multirow{2}{*}{\textbf{Stages}} & 
\multirow{2}{*}{\textbf{Data}} & 
\textbf{DAVIS} & 
\textbf{VIP} & 
\textbf{JHMDB} & 
\textbf{CamVid} & 
\textbf{HMDB} & 
\textbf{UCF} \\ 
\cmidrule(lr){5-10}
 & & & & 
$J\&F$ & 
mIoU & 
PCK & 
mIoU & 
Acc. & 
Acc. \\ 
\hline
 &  & Stage 1 & Kinetics & 62.1 & 39.0 & 46.6 & 59.9 & 52.9 & 56.9 \\ 
\checkmark & \checkmark & Stage 2 Scratch & Kinetics & 65.4 & 40.4 & 62.3 & 61.9 & 50.9 & 56.2 \\
\checkmark & \checkmark & 2-Stage FT & Kinetics & 65.7 & 41.6 & 48.5 & 61.5 & 51.1 & 59.7 \\ 
\rowcolor{cyan!15}
\checkmark & \checkmark & 2-Stage & Kinetics & 65.7 & 41.2 & 48.7 & 61.6 & 49.7 & 56.3 \\

&  & 2-Stage & Kin.+Ego4D & 62.6 & 39.2 & 46.7 & 60.1 & - & - \\ 
\checkmark &  & 2-Stage & Kin.+Ego4D & 65.5 & 41.2 & 48.6 & 62.3 & 51.4 & 59.3 \\ 
\rowcolor{cyan!15}
\checkmark & \checkmark & 2-Stage & Kin.+Ego4D & 66.3 & 42.0 & 49.0 & 62.8 & 52.1 & 58.9 \\ 
\hline
\end{tabular}
\label{tab:ablation-study}
\end{table*}

\FloatBarrier

See the Supplemental for many more ablations, including: number of memory frames, dataset size, encoder depth, decoder depth, number of training epochs, and training image resolution.

\section{Conclusion}
\label{sec:conclusion}
We introduce \frm{}, a self-supervised video frame encoder that distills dense spatial and semantic features into more compact backbones and enhances them with lightweight memory and anticipation modules. \frm{} delivers state-of-the-art performance on a range of dense video prediction tasks while maintaining a compact and efficient architecture. By aligning both patch-level and semantic representations, it offers a practical foundation for temporally consistent video understanding.

\boldheader{Limitations} We show \frm{}'s features are strong in a frozen setting but do not evaluate their performance under fine-tuning, though this is consistent with prior self-supervised work like DINO. 

\boldheader{Future work} Includes integrating \frm{} into video-language models as a drop-in backbone and extending its memory design to better encode fine-grained motion and long-term dynamics for complex video understanding.

\section{Acknowledgment}
This work is supported in part by awards ONR N00014-23-1-2383 and DARPA HR0011-23-9-0060. The views and conclusions expressed are those of the authors, and not necessarily representative of the US Government or its agencies.

{
    \small
    \bibliographystyle{abbrvnat}
    \bibliography{neurips_2025}
}

\clearpage
\appendix
\setcounter{page}{1}
\maketitle

\section{Supplementary}
\noindent{}This section is structured as follows. In \cref{sec:hyperparam-analysis}, we analyze the sensitivity of \frm{} to various hyperparameters. Specifically, we evaluate the impact of encoder depth (\cref{subsec:encoder-depth}), number of training epochs (\cref{subsec:number-of-epochs}), dataset size (\cref{subsec:dataset-size}), decoder depth (\cref{subsec:decoder-depth}), number of frames used for memory (\cref{subsec:frames}), and encoder vs. decoder features in pre-training (\cref{subsec:pretraining-datasets}) on performance. In \cref{sec:Training-details}, we outline the training methodology employed for \frm{}, including hardware setup, optimization strategy, and loss functions. Furthermore, we present our pooled region-based tracking pipeline, discuss its extension to multi-object tracking, and show qualitative examples. Finally, we describe our empirical procedure for selecting the semantic and spatial anticipation frame deltas used during training. The blue highlight in all tables indicates the selected configuration reported in the main paper.


\section{Hyperparameter Analysis} \label{sec:hyperparam-analysis}
\subsection{Impact of Encoder Depth} \label{subsec:encoder-depth}
\noindent{}Table \ref{tab:encoder-depth} shows the effect of varying encoder depth on the performance of \frm{}. For DAVIS~\cite{davis}, VIP~\cite{vip}, JHMDB~\cite{jhmdb}, and CamVid~\citet{camvid}, we observe a performance increase as the encoder depth increases from 8 to 18 layers. Notably, encoder depths of 10 and 12 both deliver strong results, with 12 being the one reported in the paper (highlighted in bold letters) due to its balance between computational efficiency and performance. While further increases in depth beyond 12 (e.g., to 14 or 18) lead to marginal gains, the increase in computational cost outweighs the slight improvement. However, for HMDB-51~\cite{hmdb51} and UCF-101~\cite{ucf101}, the trend is less clear, with results fluctuating across different depths.

\subsection{Impact of Number of Epochs} \label{subsec:number-of-epochs}
\noindent{}Table \ref{tab:number-of-epochs} highlights the sensitivity of \frm{} to the number of training epochs. For DAVIS, VIP,  JHMDB, and CamVid, there is a increase in performance as the number of epochs increases, peaking at 70 epochs. Beyond this point, performance trends vary, likely due to overfitting. For HMDB-51 and UCF-101, the trend remains unclear. This could be explained by the fact that these datasets use a semantic decoder for the current frame (refer to the Methods section of the main paper), which mimics CLIP features. As a result, they converge very quickly within a few epochs, leading to the observed fluctuations in trends.

\subsection{Impact of Dataset Size} \label{subsec:dataset-size}
\noindent{}Table \ref{tab:dataset-size} examines the effect of pretraining dataset size (percentage of Kinetics-400) on downstream task performance. For most datasets, increasing the dataset size up to 40\% results in a steady improvement in performance, with DAVIS and VIP particularly benefiting. However, beyond 60\%, the trends vary across tasks; while HMDB-51 and UCF-101 show minor gains, the other datasets exhibit less clear or negligible improvements. This suggests that although increasing the dataset size typically improves model performance, the gains diminish beyond a certain threshold, highlighting the potential need for factors like data diversity to further enhance performance.

\subsection{Impact of Decoder Depth} \label{subsec:decoder-depth}
\noindent{}Table \ref{tab:decoder-depth} evaluates the influence of decoder depth on performance. For DAVIS, VIP, JHMDB, and CamVid, a smaller decoder (fewer layers) generally achieves better results. Specifically, a decoder with depth 1 or 2 performs best, as increasing decoder depth leads to a noticeable decline in performance. This suggests that a shallow decoder is important for extracting meaningful features from the encoder. Here, the results correspond to DINO-ViT-S/8.

\subsection{Impact of Number of Frames} \label{subsec:frames}
\noindent{}Table \ref{tab:frames} examines the effect of varying the number of past frames used for memory on downstream task performance. Generally, increasing the number of frames up to 4-5 results in improved performance across all datasets, with DAVIS \cite{davis}, VIP \cite{vip}, and JHMDB \cite{jhmdb} showing the most significant gains. However, beyond this threshold, performance begins to decline. This suggests that short-term memory is more beneficial for tasks like video object segmentation and video semantic segmentation, where the relationship between past frames and the current frame remains meaningful only within a limited temporal window. After a certain point, the scene or object appearance can change drastically—an object in frame 1 might differ significantly from the same object in frame 8 but could still be similar to frames 4 or 5. This highlights the importance of selecting an optimal temporal context that balances useful continuity while avoiding outdated or irrelevant information.

\subsection{Impact of Encoder vs. Decoder Features in Pre-training} \label{subsec:pretraining-datasets}
\noindent{}Table \ref{tab:pretraining-datasets} examines the effect of different pre-training datasets on downstream task performance while comparing features extracted from the encoder and decoder. Across all datasets, the encoder features consistently outperform the decoder features, irrespective of the pre-training dataset. This suggests that the encoder captures richer temporal representations that benefit video object segmentation and video semantic segmentation tasks. The advantage likely stems from the encoder's ability to model anticipation and memory. Furthermore, adding datasets such as Ego4D and SA-V leads to marginal improvements, particularly in DAVIS, VIP, and JHMDB, indicating that additional pre-training data enhances feature generalization.

\section{Training Details}
\label{sec:Training-details}
\paragraph{Training Setup.}
We train \frm{} using a combination of raw video frames and precomputed features from DINO and CLIP. The model is optimized to predict DINO patch features and CLIP class tokens for the current frame in a sliding window of 7 RGB frames (5 past, 1 current, 1 future). DINO patch features are supervised with a mean squared error (MSE) loss, while CLIP class tokens are trained using cosine embedding loss, applied after feature normalization. Training is conducted on 8–16 NVIDIA A100 GPUs using PyTorch with the \textit{Accelerate} library\footnote{\url{https://huggingface.co/docs/transformers/en/accelerate}} for multi-GPU distributed data parallelism (DDP), gradient accumulation, and mixed precision (fp16) training via PyTorch AMP. This setup enables efficient large-batch training within memory constraints while improving computational throughput. 

We use the Adam optimizer with a learning rate of $1\mathrm{e}{-4}$ and weight decay of $1\mathrm{e}{-4}$, following a cosine annealing schedule with linear warm-up to ensure stable convergence. The training batch size is 8 per GPU, and training runs for up to 70 epochs with early stopping after 5 epochs of no improvement in validation loss. Each training sample is paired with precomputed DINO~\cite{dinov1,dinov2} and CLIP~\cite{clip} features. We train on two large-scale video datasets: Kinetics~\cite{kinetics} and Ego4D~\cite{Grauman2021Ego4DAT}, using 80\% of the data for training and 20\% for validation. Data loading is parallelized with 16 workers and pin-memory enabled. All training metrics, including loss breakdowns and validation performance, are logged with \texttt{wandb}. The training pipeline is designed to scale efficiently across large distributed clusters, with built-in support for mixed precision and dynamic learning rate adjustment.


\begin{table*}[ht]
\centering
\small
\renewcommand{\arraystretch}{1.2}
\begin{tabular}{c c c c c c}
\hline
\multirow{2}{*}{\textbf{Encoder Depth}} & 
\textbf{\small DAVIS} & 
\textbf{\small VIP} & 
\textbf{\small JHMDB} & 
\textbf{\small HMDB-51} & 
\textbf{\small UCF-101} \\ 
\cmidrule(lr){2-6}
\textbf{} & 
$J\&F_m$ & 
mIoU & 
PCK@0.1 & 
\% & 
\% \\ 
\hline
8 & 63.5 & 41.7 & 47.3 & 53.3 & 58.9 \\ 
10 & 64.9 & 41.8 & 48.6 & 53.1 & 59.4 \\ 
\rowcolor{cyan!15}
12 & 66.3 & 42.0 & 49.0 & 52.1 & 58.5 \\ 
14 & 66.7 & 42.1 & 49.0 & 51.3 & 59.2 \\ 
18 & 66.9 & 42.3 & 49.1 & 51.2 & 58.2 \\ 
\hline
\end{tabular}
\caption{
\textbf{Impact of Encoder Depth.}
We evaluate the impact of increasing encoder depth. Our final evaluations use Encoder Depth = 12}
\label{tab:encoder-depth}
\end{table*}

\begin{table*}[ht]
\centering
\small
\renewcommand{\arraystretch}{1.2}
\begin{tabular}{c c c c c c c}
\hline
\multirow{2}{*}{\textbf{Number of Epochs}} & 
\textbf{\small DAVIS} & 
\textbf{\small VIP} & 
\textbf{\small JHMDB} & 
\textbf{\small CamVid} & 
\textbf{\small HMDB-51} & 
\textbf{\small UCF-101} \\ 
\cmidrule(lr){2-7}
\textbf{} & 
$J\&F_m$ & 
mIoU & 
PCK@0.1 & 
mIoU & 
\% & 
\% \\ 
\hline
20 & 31.3 & 19.3 & 18.2 & 20.5 & 52.7 & 56.5 \\ 
40 & 55.4 & 33.3 & 35.6 & 30.4 & 52.2 & 57.7 \\ 
60 & 65.7 & 42.1 & 48.4 & 58.7 & 52.5 & 59.2 \\ 
\rowcolor{cyan!15}
70 & 66.3 & 42.0 & 49.0 & 62.8 & 52.1 & 58.5 \\ 
90 & 66.1 & 42.0 & 48.7 & 62.9 & 50.2 & 58.6 \\ 
150 & 65.2 & 41.7 & 48.6 & 61.2 & 51.5 & 55.7 \\ 
\hline
\end{tabular}
\caption{
\textbf{Impact of Number of Epochs.}
We evaluate the effect of training for different numbers of epochs. Our final evaluations correspond to 70 epochs.
}
\label{tab:number-of-epochs}
\end{table*}


\begin{table*}[ht]
\centering
\small
\renewcommand{\arraystretch}{1.2}
\begin{tabular}{c c c c c}
\hline
\multirow{2}{*}{\textbf{Kinetics (\%)}} & 
\textbf{\small DAVIS} & 
\textbf{\small VIP} & 
\textbf{\small JHMDB} \\ 
\cmidrule(lr){2-4}
& $J\&F_m$ & mIoU & PCK@0.1 \\ 
\hline
20\% & 55.7 & 36.8 & 41.6 \\ 
30\% & 63.2 & 40.3 & 45.1 \\
\rowcolor{cyan!15}
40\% & 65.6 & 41.2 & 48.7 \\ 
70\% & 65.2 & 41.4 & 48.8 \\ 
80\% & 65.3 & 41.0 & 48.5 \\ 
100\% & 64.9 & 41.2 & 48.3 \\ 
\hline
\end{tabular}
\caption{
\textbf{Impact of Dataset Size.}
We evaluate the effect of varying the percentage of the Kinetics-400 dataset used for pretraining. Our final evaluations use 40\% of the dataset.
}
\label{tab:dataset-size}
\end{table*}


\begin{table*}[ht]
\centering
\small
\renewcommand{\arraystretch}{1.2}
\begin{tabular}{c c c c c}
\hline
\multirow{2}{*}{\textbf{Resolution}} & 
\textbf{\small DAVIS} & 
\textbf{\small VIP} & 
\textbf{\small JHMDB} & 
\textbf{\small CamVid} \\ 
\cmidrule(lr){2-5}
\textbf{} & 
$J\&F_m$ & 
mIoU & 
PCK@0.1 & 
mIoU \\ 
\hline
224$\times$224 & 63.5 & 38.5 & 48.5 & 61.0 \\ 
320$\times$320 & 64.5 & 40.0 & 48.5 & 62.1 \\ 
\rowcolor{cyan!15}
400$\times$400 & 66.3 & 42.0 & 49.0 & 62.8 \\ 
512$\times$512 & 66.5 & 42.5 & 49.0 & 62.8 \\ 
720$\times$720 & 66.5 & 42.0 & 49.0 & 62.7 \\ 
\hline
\end{tabular}
\caption{
\textbf{Impact of Image Resolution.}
We evaluate the effect of varying the input image resolution on performance across different datasets. Performance generally increases with resolution before plateauing at 512$\times$512.
}
\label{tab:resolution_metrics}
\end{table*}

\begin{table*}[ht]
\centering
\small
\renewcommand{\arraystretch}{1.2}
\begin{tabular}{c c c c c c c}
\hline
\multirow{2}{*}{\textbf{Number of Frames}} & 
\textbf{\small DAVIS} & 
\textbf{\small VIP} & 
\textbf{\small JHMDB} & 
\textbf{\small CamVid} & 
\textbf{\small UCF101} &
\textbf{\small HMDB101} \\ 
\cmidrule(lr){2-7}
\textbf{} & 
$J\&F_m$ & 
mIoU & 
PCK@0.1 & 
mIoU &
mIoU &
Accuracy \\ 
\hline
1 & 61.3 & 38.4 & 46.1 & 59.2 & 60.2 & 53.3 \\ 
2 & 62.1 & 40.7 & 46.9 & 60.3 & 58.9 & 50.6 \\ 
3 & 63.6 & 41.0 & 47.8 & 62.1 & 59.2 & 51.2 \\ 
4 & 65.7 & 42.2 & 48.3 & 63.2 & 56.3 & 50.4 \\
\rowcolor{cyan!15}
5 & 66.3 & 42.4 & 49.0 & 62.8 & 58.5 & 52.1 \\
6 & 63.8 & 39.5 & 48.6 & 61.3 & 56.4 & 53.1 \\
7 & 63.9 & 39.4 & 43.2 & 61.2 & 58.6 & 52.2 \\
\hline
\end{tabular}
\caption{
\textbf{Impact of Number of Memory Frames.}
We evaluate the effect of varying the number of past frames for memory on performance across different datasets. Performance generally peaks with 4-5 frames, then begins to decline with more frames, likely due to increased complexity of motion.
}
\label{tab:frames}
\end{table*}

\begin{table*}[ht]
\centering
\small
\renewcommand{\arraystretch}{1.2}
\begin{tabular}{c c c c c c c}
\hline
\multirow{2}{*}{\textbf{Method}} & 
\multirow{2}{*}{\textbf{Backbone}} & 
\multirow{2}{*}{\textbf{Pre-training Dataset}} & 
\textbf{\small DAVIS} & 
\textbf{\small VIP} & 
\textbf{\small JHMDB} & 
\textbf{\small CamVID} \\ 
\cmidrule(lr){4-7}
\textbf{} & 
\textbf{} & 
\textbf{} & 
$J\&F_m$ & 
mIoU & 
PCK@0.1 & 
mIoU \\ 
\hline
\rowcolor{cyan!15}

\frm{} (encoder) & ViT S/16 & Kinetics & 65.7 & 41.2 & 48.7 & 62.1 \\
\frm{} (decoder) & ViT S/16 & Kinetics & 62.1 & 39.4 & 46.3 & 59.2 \\
\rowcolor{cyan!15}

\frm{} (encoder) & ViT S/16 & Kinetics + Ego4D & 66.3 & 42.0 & 49.0 & 62.4 \\
\frm{} (decoder) & ViT S/16 & Kinetics + Ego4D & 62.4 & 39.2 & 46.1 & 59.4 \\
\rowcolor{cyan!15}

\frm{} (encoder) & ViT S/16 & Kinetics + Ego4D + SA-V & 65.8 & 42.1 & 48.7 & 62.8 \\
\frm{} (decoder) & ViT S/16 & Kinetics + Ego4D + SA-V & 62.6 & 38.9 & 46.7 & 58.8 \\
\hline
\end{tabular}
\caption{
\textbf{Impact of Encoder vs. Decoder Features in Model Performance.} 
We analyze the effect of encoder versus decoder features for evaluation. Across all pre-training settings, features extracted from the encoder consistently yield better performance than those from the decoder, highlighting the benefits of anticipation and memory in video understanding.
}

\label{tab:pretraining-datasets}
\end{table*}

\begin{table*}[ht]
\centering
\small
\renewcommand{\arraystretch}{1.2}
\begin{tabular}{c c c c c c c c}
\hline
\multirow{2}{*}{\textbf{Teacher Net}} & 
\textbf{\small DAVIS} & 
\textbf{\small VIP} & 
\textbf{\small JHMDB} & 
\multicolumn{2}{c}{\textbf{CamVid (mIoU)}} & 
\textbf{\small HMDB-51} & 
\textbf{\small UCF-101} \\ 
\cmidrule(lr){5-6}
\textbf{} & 
\textbf{} & 
\textbf{} & 
PCK@0.1 & 
Current & 
Future & 
Accuracy & 
Accuracy \\ 
\hline
\rowcolor{cyan!15}
DINO+CLIP & 74.4 & 48.2 & 64.3/86.0 & 62.6 & 54.0 & 76.1 & 69.7 \\
DINO & 74.1 & 48.1 & 64.4/86.0 & 62.5 & 54.1 & 56.4 & 52.7 \\
\hline
\end{tabular}
\caption{
\textbf{Abalating the teacher network.} 
Comparison between DINO+CLIP and standalone DINO as teacher networks across multiple benchmarks.
}
\label{tab:dino-only}
\end{table*}

\begin{table*}[ht]
\centering
\small
\renewcommand{\arraystretch}{1.2}
\begin{tabular}{c c c c c}
\hline
\multirow{2}{*}{\textbf{Method}} & 
\multicolumn{2}{c}{\textbf{UCF-101}} & 
\multicolumn{2}{c}{\textbf{HMDB-51}} \\ 
\cmidrule(lr){2-3} \cmidrule(lr){4-5}
 & Zero-shot & Linear & Zero-shot & Linear \\ 
\hline
\frm{} (MSE) & 32.4 & 50.4 & 39.2 & 68.9 \\
\frm{} (Cosine Embedding Loss) & 55.4 & 67.9 & 48.1 & 75.1 \\
\frm{} (Cosine + Mean Scaling) & 57.2 & 68.8 & 50.3 & 75.6 \\
\rowcolor{cyan!15}
\frm{} (Cosine + Mean Scaling + Cosine Annealing) & 58.5 & 69.7 & 52.1 & 76.1 \\
\frm{} (+ Loss Dropout) & 58.3 & 69.9 & 51.9 & 75.8 \\
\hline
\end{tabular}
\caption{
\textbf{Comparison of Loss Functions for Video Classification.} 
Performance of \frm{} using different loss strategies when learning CLIP's global semantic features (CLS token). .
}
\label{tab:frame-loss-comparison}
\end{table*}

Our ablation study \ref{tab:frame-loss-comparison} evaluates different loss function strategies for optimizing \frm{}'s alignment with CLIP's class token. We begin with a baseline Mean Squared Error (MSE) approach, which yields modest results (32.4\% zero-shot accuracy on UCF-101 and 39.2\% on HMDB-51). Switching to Cosine Embedding Loss substantially improves performance (55.4\% and 48.1\% zero-shot accuracy respectively) by directly optimizing for directional similarity between embedding vectors. Adding Mean Scaling further enhances results (57.2\% and 50.3\%) by normalizing loss components to prevent one component from dominating the gradient updates, ensuring balanced learning across both semantic and spatial features. The combination of Cosine Embedding Loss with Mean Scaling and Cosine Annealing learning rate scheduling achieves the best performance (58.5\% and 52.1\%), as the warm restarts help escape local minima during optimization. Our exploration of techniques like Loss Dropout on top of the final version (temporarily disabling one loss component) don't yield further improvements, suggesting our approach with cosine annealing already achieves near-optimal alignment between \frm{}'s class token representations and CLIP's semantic space. We have skipped including the variations in how well patch tokens align with DINO's patch tokens, since the impact was minimal.

\begin{table*}[ht]
\centering
\small
\renewcommand{\arraystretch}{1.2}
\begin{tabular}{c c c c ccc cc}
\hline
\multirow{2}{*}{\textbf{Backbone}} & \multirow{2}{*}{\textbf{Dataset}} & \multicolumn{3}{c}{\textbf{DAVIS}} & \textbf{VIP} & \textbf{JHMDB} \\ 
\cmidrule(lr){3-5} \cmidrule(lr){6-6} \cmidrule(lr){7-7}
 &  & $J\&F_m$ & $J_m$ & $F_m$ & mIoU & PCK@0.1 / PCK@0.2 \\ 
\hline
ViT-S/16 & Kinetics & 65.7 & 62.1 & 69.2 & 41.2 & 48.7 / 79.2 \\
ViT-S/16 & Kinetics+Ego4D & 66.3 & 62.9 & 69.8 & 42.0 & 49.0 / 79.3 \\
ViT-S/16 & Kinetics+Ego4D+SA-V & 65.8 & 62.7 & 68.9 & 42.1 & 48.7 / 78.9 \\
\hline
ViT-S/8 & Kinetics & 73.2 & 69.5 & 77.0 & 47.9 & 64.1 / 85.9 \\
ViT-S/8 & Kinetics+Ego4D & 74.4 & 69.9 & 79.0 & 48.2 & 64.3 / 86.0 \\
ViT-S/8 & Kinetics+Ego4D+SA-V & 74.2 & 69.7 & 79.0 & 48.4 & 64.2 / 86.0 \\
\hline
\end{tabular}
\caption{
\textbf{Data Ablation Study for \(\textbf{\frm{}}\).} 
Performance comparison using different training datasets across ViT-S/16 and ViT-S/8 architectures.
}
\label{tab:data_ablation_table}
\end{table*}


\begin{table*}[ht]
\centering
\small
\renewcommand{\arraystretch}{1.2}
\begin{tabular}{c c c c c c}
\hline
\multirow{2}{*}{\textbf{Decoder Depth}} & 
\textbf{\small DAVIS} & 
\textbf{\small VIP} & 
\textbf{\small JHMDB} & 
\textbf{\small CamVid} & 
\textbf{\small VSPW} \\ 
\cmidrule(lr){2-6}
& $J\&F_m$ & mIoU & PCK@0.1 & mIoU & mIoU \\ 
\hline
\rowcolor{cyan!15}
1 & 66.3 & 42.0 & \textbf{49.0} & \textbf{62.8} & \textbf{38.9} \\ 
2 & \textbf{66.6} & \textbf{42.2} & 48.9 & 61.4 & 38.1 \\ 
3 & 63.7 & 41.9 & 48.1 & 60.3 & 37.3 \\ 
5 & 62.5 & 41.6 & 45.3 & 59.1 & 37.0 \\ 
\hline
\end{tabular}
\caption{
\textbf{Impact of Decoder Depth.} Increasing the decoder depth beyond 2 layers degrades performance across all datasets. Our final model uses a shallow decoder with depth = 1.
}
\label{tab:decoder-depth}
\end{table*}

\begin{table*}[ht]
\centering
\small
\renewcommand{\arraystretch}{1.2}
\begin{tabular}{c c c c c c}
\hline
\multirow{2}{*}{\textbf{Method}} & 
\textbf{\small DAVIS} & 
\textbf{\small VIP} & 
\textbf{\small JHMDB} & 
\textbf{\small CamVid} & 
\textbf{\small VSPW} \\ 
\cmidrule(lr){2-6}
\textbf{} & 
$J\&F_m$ & 
mIoU & 
PCK@0.1 & 
mIoU & 
mIoU \\ 
\hline
\frm{} (w/o memory) & 62.6 & 39.2 & 46.7 & 60.1 & 37.3 \\ 
\frm{}  (w/o anticipation) & 65.5 & 41.2 & 48.6 & 62.3 & 38.0 \\ 
\rowcolor{cyan!15}
\frm{} (w/ memory + w/ anticipation) & 66.3 & 42.0 & 49.0 & 62.8 & 38.9 \\ 
\hline
\end{tabular}
\caption{
\textbf{Impact of Memory and Anticipation Components.}
We evaluate the impact of memory and anticipation on \frm{} performance across different datasets when pre-trained with Kinetics and Ego4d.
}
\label{tab:temporal-ablation}
\end{table*}


\begin{table*}[ht]
\centering
\small
\renewcommand{\arraystretch}{1.2}
\begin{tabular}{l c c c c}
\hline
\multirow{2}{*}{\textbf{Model}} & 
\multicolumn{2}{c}{\textbf{Zero-Shot (\%)}} & 
\multicolumn{2}{c}{\textbf{Linear (\%)}} \\ 
\cmidrule(lr){2-3} \cmidrule(lr){4-5}
& \textbf{HMDB-51} & \textbf{UCF-101} & \textbf{HMDB-51} & \textbf{UCF-101} \\ 
\hline
CLIP ViT-L/14 & \textbf{55.9} & \textbf{60.6} & 78.1 & 71.3 \\ 
\frm{} ViT-S/16 & 55.1 & 58.5 & 76.1 & 71.1 \\ 
\frm{} ViT-B/16 & 55.7 & 59.6 & \textbf{78.5} & \textbf{72.2} \\ 
\frm{} ViT-L/14 & 55.3 & 59.9 & 76.3 & 71.6 \\ 
\hline
\end{tabular}
\caption{
\textbf{Video Activity Classification Comparison.}
We report zero-shot and linear classification accuracy on HMDB-51 and UCF-101. \frm{} performs competitively with CLIP (ViT-L/14).}
\label{tab:clip_classification_comparison}
\end{table*}

\begin{table}[t]
    \centering
    \small
    \setlength{\tabcolsep}{4pt}
    \renewcommand{\arraystretch}{1.15}
    \resizebox{0.6\linewidth}{!}{%
        \begin{tabular}{lcc}
            \toprule
            \textbf{Model} & \textbf{CamVid (mIoU)} & \textbf{VSPW (mIoU)} \\
            \midrule
            PIDNet~\cite{PIDNet} & \textbf{84.6} & -- \\
            DVIS++~\cite{dvis++} & -- & \textbf{63.8} \\
            DINOv2 ViT-L/14 & 74.1 & 46.5 \\
            \frm{} ViT-L/14 & 77.9 & 52.3 \\
            \bottomrule
        \end{tabular}
    }
    \caption{
        \textbf{Semantic Segmentation Results.}
        \frm{} approaches state-of-the-art performance (PIDNet~\cite{PIDNet} for CamVid and DVIS++~\cite{dvis++} for VSPW) using a simple U-Net decoder.
    }
    \label{tab:segmentation_comparison}
\end{table}

\begin{table*}[ht]
\centering
\small
\renewcommand{\arraystretch}{1.2}
\setlength{\tabcolsep}{6pt}
\begin{tabular}{l l l}
\toprule
\textbf{Dataset} & \textbf{License} & \textbf{Notes} \\
\midrule
DAVIS 2017 & CC BY-NC-SA 4.0 & Academic use only, non-commercial \\
Kinetics-400 & DeepMind Terms of Use & Freely available for research \\
Ego4D & Ego4D Data Use Agreement & Requires signed agreement for access \\
UCF-101 & Custom (UCF License) & Research only; redistribution prohibited \\
HMDB-51 & Custom (HMDB License) & Academic and research use only \\
JHMDB & Follows HMDB terms & Derived from HMDB, same constraints \\
VIP (Video Instance Parsing) & CC BY-NC-SA 4.0 & Research use, attribution required \\
CamVid & Custom (CamVid Terms) & Academic use with citation \\
VSPW & CC BY-NC-SA 4.0 & Research permitted, commercial use restricted \\
\bottomrule
\end{tabular}
\caption{\textbf{Licensing Information for Datasets Used in Our Work.} All the datasets are standard academic datasets and are publicly available for academic research.}
\label{tab:dataset_licenses}
\end{table*}

\subsection*{Broader Impact}
This work introduces \frm{}, a self-supervised video frame encoder that distills dense spatial and semantic features from pretrained image models, and augments them with lightweight temporal modeling to produce temporally consistent video representations. By leveraging existing image-based models (e.g., DINO, CLIP) rather than training from scratch on large-scale video datasets, our approach significantly reduces the computational cost and data requirements typically associated with video model training. The modular and efficient nature of \frm{} promotes accessibility and scalability in domains where dense video prediction is valuable but labeled data is scarce—such as in education, scientific research, healthcare, and environmental monitoring. Additionally, by lowering the barrier to entry for high-quality video representation learning, our method may contribute to the broader democratization of video AI technologies.

However, since \frm{} builds upon pretrained image models, it may inherit biases present in those foundation models. This can affect fairness, especially in downstream applications involving human-centric or safety-critical scenarios. Moreover, mispredictions—such as incorrect temporal alignment or false localization—could lead to cascading failures in safety-critical systems like autonomous navigation, medical video analysis, or security applications. To mitigate these risks, we encourage responsible deployment practices, including human-in-the-loop systems, clear documentation of limitations, and proactive auditing of downstream use cases. As a foundational model, the societal impact of \frm{} will depend significantly on how it is integrated into applied systems.

\begin{table*}[ht]
\centering
\small
\renewcommand{\arraystretch}{1.2}
\setlength{\tabcolsep}{1pt}  
\resizebox{0.85\textwidth}{!}{%
\begin{tabular}{l c c c c}
\toprule
\textbf{Training Set} & \textbf{DAVIS ($J\&F_m$)} & \textbf{VIP (mIoU)} & \textbf{JHMDB (PCK@0.1)} & \textbf{CamVid (mIoU)} \\
\midrule
Kinetics-400 (80K) + Ego4D Subset 1 & 66.1 & 42.3 & 48.7 & 62.4 \\
Kinetics-400 (80K) + Ego4D Subset 2 & 66.4 & 42.1 & 48.9 & 63.0 \\
Kinetics-400 (80K) + Ego4D Subset 3 & 66.0 & 42.2 & 48.7 & 62.7 \\
Kinetics-400 (80K) + Ego4D Subset 4 & 65.8 & 42.0 & 48.5 & 62.5 \\
\rowcolor{cyan!15}
Kinetics-400 (80K) + Ego4D (700 videos) & 66.3 & 42.4 & 49.0 & 62.8 \\
Full Kinetics-400 + Full Ego4D & 66.5 & 42.6 & 48.9 & 63.1 \\
\bottomrule
\end{tabular}
}
\caption{
\textbf{Impact of Training Data Composition.}
We compare \frm{} performance when trained on various Ego4D subsets (combined with Kinetics-400). Subset results are stable, with minor variation across tasks. Final numbers reported in the paper use 80K Kinetics-400 videos and 700 Ego4D videos (highlighted in blue). Slight performance gains are observed with full-scale pretraining, while some subsets match or even exceed full training on specific tasks.
}
\label{tab:training_ablation}
\end{table*}

\begin{table*}[ht]
\centering
\small
\renewcommand{\arraystretch}{1.2}
\setlength{\tabcolsep}{6pt}
\resizebox{0.95\textwidth}{!}{%
\begin{tabular}{l c c c c}
\toprule
\textbf{Model} & \textbf{Params (M)} & \textbf{DAVIS ($J\&F_m$)} & \textbf{VIP (mIoU)} & \textbf{JHMDB (PCK@0.1)} \\
\midrule
DINO ViT-S/16        & 21    & 61.8 & 36.2 & 45.6 \\
\textbf{FRAME ViT-S/16} & 38    & 66.3 & 42.0 & 49.0 \\
DINO ViT-S/8         & 21    & 69.9 & 39.5 & 56.5 \\
\textbf{FRAME ViT-S/8}  & 38    & 74.4 & 48.2 & 64.3 \\
DINO ViT-B/16        & 85    & 63.2 & 39.1 & 48.2 \\
\textbf{FRAME ViT-B/16} & 101   & 68.2 & 42.7 & 49.4 \\
DINO ViT-B/8         & 85    & 72.3 & 44.6 & 58.9 \\
\textbf{FRAME ViT-B/8}  & 101   & 75.1 & 49.7 & 65.7 \\
DINOv2 ViT-L/14      & 300   & 64.6 & 43.4 & 48.3 \\
\textbf{FRAME ViT-L/14} & 109   & 67.9 & 45.1 & 51.3 \\
\bottomrule
\end{tabular}
}
\caption{
\textbf{Model Scale, Runtime, and Performance Comparison.}
We report parameter count and performance on DAVIS~\cite{davis}, VIP~\cite{vip}, and JHMDB~\cite{jhmdb} for DINO (v1/v2) and \frm{} across various ViT backbones and patch sizes. \frm{} consistently improves over the corresponding DINO baseline at each scale, and shows strong gains with smaller models like ViT-S/8. This highlights (1) the benefit of moving from ViT-S to ViT-B, (2) FRAME’s improvement over DINO at every patch resolution, and (3) a favorable accuracy-vs-compute trade-off for FRAME models.
}
\label{tab:model_comparison}
\end{table*}

\begin{figure}[t]
    \centering
    \includegraphics[width=\linewidth]{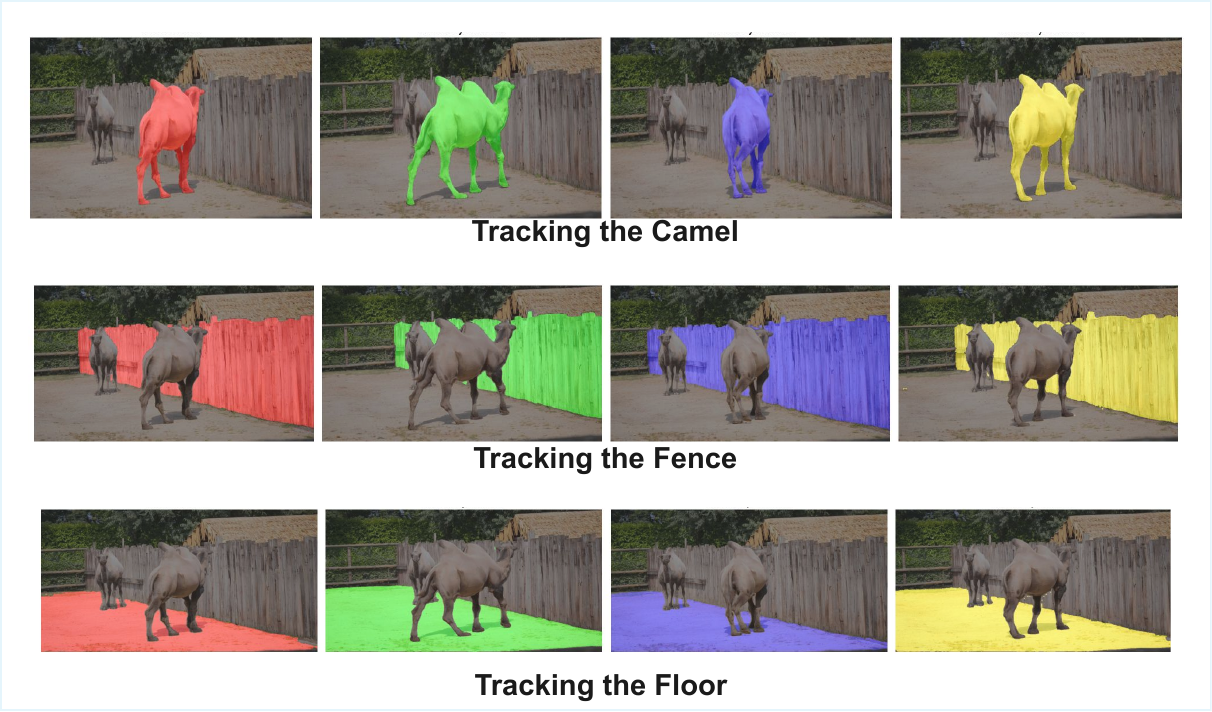}
    \caption{{\bf Region-based tracking with \frm{}.}  We track different objects—eg. camel, fence, and floor—across time using FRAME-pooled region features. Each row shows the temporal propagation of a single object, demonstrating the ability of \frm{} to support accurate and consistent tracking of diverse regions.}
    \label{fig:qualitative_examples_regions}
    \vspace{-12pt}
\end{figure}

\paragraph{Pooled Region-Based Tracking.}
To evaluate the temporal consistency of \frm{} features, we adopt a pooled region-based tracking approach inspired by prior region-centric pipelines~\cite{dinosam, relocate}. For each video frame, we extract spatially dense patch tokens using the \frm{} encoder. We then apply an off-the-shelf segmentation model such as DINO-SAM~\cite{dinosam} to generate region proposals. For each region, we compute a region-level feature by mean-pooling the \frm{} patch tokens corresponding to the region's mask. To propagate object identity across time, we perform nearest-neighbor matching of these pooled region features between consecutive frames using cosine similarity. Tracking is initialized from the annotated mask in the first frame by extracting its pooled feature and then iteratively matching it to regions in subsequent frames. Compared to patch-level correspondence, this region-centric formulation could be more robust to intra-object appearance variations and local noise. Moreover, \frm{} features exhibit significantly better temporal consistency than DINO ( as presented in the main paper Table 3), enabling more accurate region matching across frames (Fig . \ref{fig:qualitative_examples_regions}). We evaluate this tracking pipeline on DAVIS~\cite{davis}, measuring performance using the standard $J\&F_m$ metric. \frm{} outperforms DINO variants in this setup, while being close to SAM 2\cite{sam2}.

\paragraph{Multi-Object Tracking via Region Pools.}
A key strength of this pooled region-based approach is its natural extension to \emph{multi-object tracking}. Unlike models such as SAM2, which require iterative prompting or segmentation for each object in each frame—leading to high computational and memory overhead—our approach could potentially all objects simultaneously using pooled features derived from a single forward pass of \frm{}. Specifically, we initialize pooled region embeddings for all masks in the annotated frame and independently propagate them forward through the video using per-region cosine similarity. This allows us to track multiple objects in parallel without re-encoding the video or repeatedly invoking a segmentation model. As a result, our method could enable efficient many-object tracking and segmentation at scale, making it especially suitable for long videos, memory-constrained settings, or applications involving dense object populations. We view this as a promising future direction for leveraging generic video encoders like \frm{} for scalable open-world video object tracking.

\subsection{}{ Frame Delta Selection for Anticipation Decoders}
To determine appropriate frame offsets (\textit{frame deltas}) for the semantic and spatial anticipation decoders in \frm{}, we conducted an empirical analysis of feature variability across time. The goal was to select temporal distances that induce meaningful representation change—sufficient to enable predictive learning—without introducing excessive noise or instability due to long-term drift or scene discontinuities.
\paragraph{Setup.}
We evaluated temporal feature variability on a subset of 200 videos sampled from the Kinetics-400~\cite{kinetics}, DAVIS \cite{davis} datasets. For each video, we extracted a sequence of frames and computed:
\begin{itemize}
    \item Global features using CLIP's ViT-B/32 class token.
    \item Local features using DINO's ViT-S/16 patch tokens.
\end{itemize}
We then measured the feature change between frame $F_t$ and $F_{t+k}$ using cosine distance. For CLIP, we computed cosine distance between $\mathbf{c}^{t}_{cls}$ and $\mathbf{c}^{t+k}_{cls}$; for DINO, we averaged cosine distances between corresponding patch tokens across the frame.

\paragraph{Results.}
Our analysis reveal that CLIP global features exhibit relatively low temporal variation, with cosine distances increasing gradually across time and DINO patch features changed more rapidly, reflecting the fine-grained and localized nature of DINO representations.

\paragraph{Choice of Deltas.}
Based on this analysis, we selected a frame delta of 4 for the semantic anticipator (CLIP-like) and a delta of 2 for the spatial anticipator (DINO-like). These deltas were chosen to balance the temporal predictiveness of each target with the model's learning capacity: they represent sufficient future context to encourage meaningful anticipation while avoiding overburdening the model with distant and unstable predictions. This design choice is further validated by downstream performance, where these deltas yield strong results across video segmentation and tracking tasks.

\end{document}